\documentclass[letterpaper, 10 pt, conference]{ieeeconf}  
\IEEEoverridecommandlockouts                              

\usepackage{graphics}    
\usepackage{times}       
\usepackage{amsmath}     
\usepackage{amssymb}     
\usepackage{graphicx}
\usepackage{tabularx}
\usepackage{booktabs}
\usepackage{multirow}

\usepackage[separate-uncertainty=true,number-unit-product=\ ]{siunitx}%
\usepackage{subcaption}
\usepackage{tabularx}
\usepackage{caption}
\usepackage[hidelinks]{hyperref} 
\usepackage{makecell}

\usepackage[rgb]{xcolor}
\definecolor{mutedorange}{RGB}{230, 120, 60} 

\usepackage{flushend}

\usepackage[font=small]{caption}

\def\figref#1{Fig.~\ref{#1}}

\def\eqref#1{Eq.~(\ref{#1})}

\newcommand{\policy}{\pi}
\newcommand{\velocity}{v}
\renewcommand{\angle}{\omega}
\newcommand{\grad}{\vec{g}}
\newcommand{\state}{s}
\newcommand{\statet}{\state_t}

\newenvironment{hypothesesenum}{
	
	\begin{enumerate}
	}{
	\end{enumerate}
}

\title{\LARGE \bf Immersive  Explainability: Visualizing Robot Navigation Decisions\\through XAI Semantic Scene Projections in Virtual Reality}

\author{Jorge de Heuvel \and Sebastian Müller \and Marlene Wessels \and Aftab Akhtar \and Christian Bauckhage \and Maren Bennewitz
  \thanks{
   All authors except M. Wessels are with the University of Bonn, Germany. 
   M. Bennewitz, C. Bauckhage, J. de Heuvel, and S. Müller are additionally with the Lamarr Institute for Machine Learning and Artificial Intelligence. 
   M. Bennewitz and J. de Heuvel are also with the Center for Robotics, Bonn, Germany. 
   M. Wessels is with the University of Mainz, Germany.
   C. Bauckhage is additionally with the Fraunhofer Institute for Intelligent Analysis and Information Systems IAIS, Sankt Augustin, Germany.
   This work has been partially funded by the BMBF, grant No. 16KIS1949 and within the Robotics Institute Germany, grant No. 16ME0999, and by the Federal Ministry of Education and Research of Germany and the state of North-Rhine Westphalia as part of the Lamarr Institute for Machine Learning and Artificial Intelligence, LAMARR22B.
   Generative AI tools contributed to minor language improvements.
   }
}

\begin{document}
\maketitle
\thispagestyle{empty} 
\pagestyle{empty}

\begin{abstract} 
End-to-end robot policies achieve high performance through neural networks trained via reinforcement learning (RL).
Yet, their black box nature and abstract reasoning pose challenges for human-robot interaction (HRI), because humans may experience difficulty in understanding and predicting the robot's navigation decisions, hindering trust development. 
We present a virtual reality (VR) interface that visualizes explainable AI (XAI) outputs and the robot’s lidar perception to support intuitive interpretation of RL-based navigation behavior.
By visually highlighting objects based on their attribution scores, the interface grounds abstract policy explanations in the scene context.
This XAI visualization bridges the gap between obscure numerical XAI attribution scores and a human-centric semantic level of explanation.
A within-subjects study with 24 participants evaluated the effectiveness of our interface for four visualization conditions combining XAI and lidar.
Participants ranked scene objects across navigation scenarios based on their importance to the robot, followed by a questionnaire assessing subjective understanding and predictability.
Results show that semantic projection of attributions significantly enhances non-expert users’ objective understanding and subjective awareness of robot behavior.
In addition, lidar visualization further improves perceived predictability, underscoring the value of integrating XAI and sensor  for transparent, trustworthy HRI.
\end{abstract}

\vspace*{-0.5em}
\section{Introduction}
\begin{figure}[t]
	\centering
	\includegraphics[width=1\linewidth]{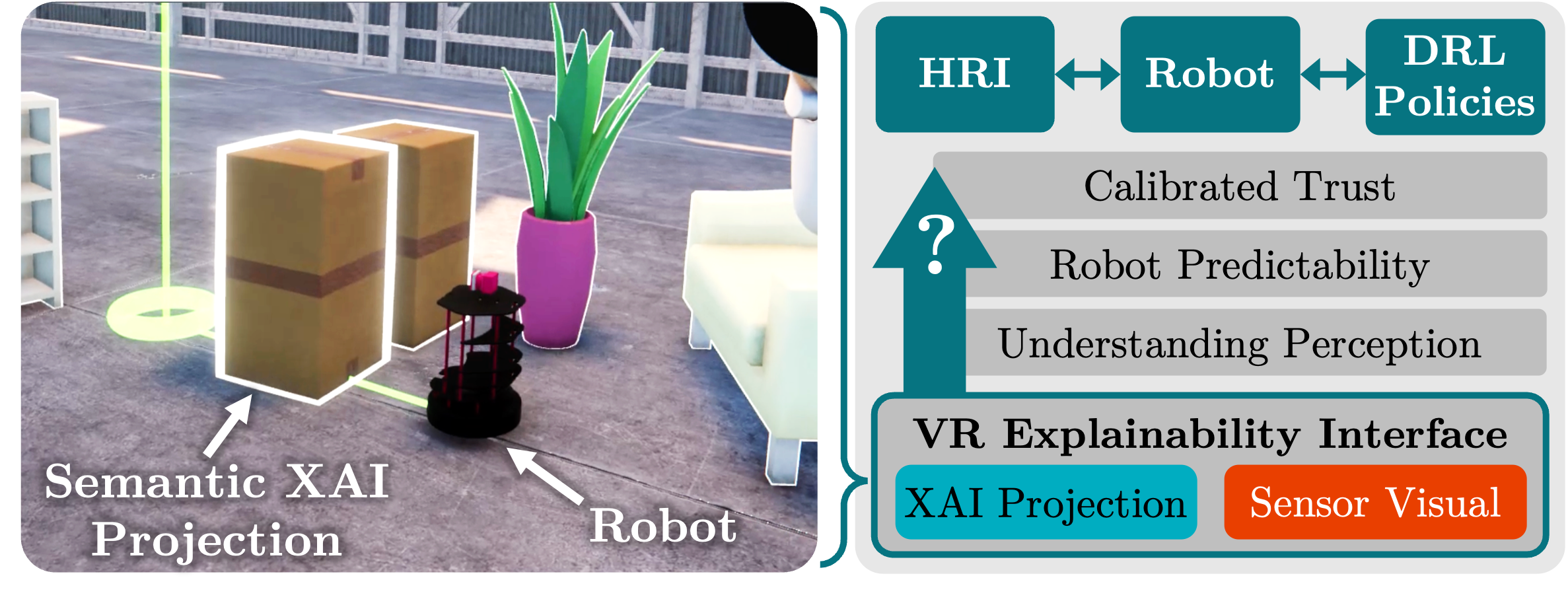}
	\caption{
        Our immersive VR explainability interface communicates XAI attributions and sensor perception of an RL robot navigation policy to non-expert users, by grounding them in the object semantics of the scene.
        Objects that are important to the policy are highlighted using a glowing outline.
        A better perception understanding in combination with the user's perceived ability predict to the robot can lead to calibrated trust towards the robot.
\label{fig:motivation}}
\end{figure}

\vspace*{-0.25em}
\label{sec:intro}
Human-robot interaction (HRI) increasingly relies on high-performance robot policies driven by deep reinforcement learning (RL), enabling robots to navigate complex and human environments with remarkable autonomy. 
However, the decision-making processes behind these policies often remain intransparent to end-users because they depend on neural networks that are effectively black boxes~\cite{puiutta_explainable_2020-1}. 
This is further compounded by the “perceptual belief problem” ~\cite{thellman_perceptual_2021} that arises from people’s difficulty in understanding what robots know about the shared environment, e.g., due to limited familiarity with robotic sensing capabilities.
The lack of understanding impedes robot predictability by the user which can impact user trust, as illustrated in \figref{fig:motivation}.

Within this context, explainable artificial intelligence (XAI) techniques have been explored to make robot decisions more comprehensible and interpretable~\cite{he_explainable_2021, edmonds_tale_2019}.
A recent study showed that the effectiveness of XAI methods across different applications varies significantly~\cite{rong_humancentered_2024}, underlining the need for more user-focused evaluations of how XAI explanations are conveyed.
This is especially challenging for the continuous, dynamic decision-making process of a navigating robot.
The question arises as to how non-expert users can effectively and intuitively understand both the robot’s perceptual capabilities and the explanations generated by XAI methods~\cite{halilovic_robot_2024, hald_error_2021}.

Therefore, we propose an immersive virtual reality (VR) interface that integrates two key elements for novice users: a clear visualization of a) the robot’s sensor data and b) the contextual XAI outputs.\footnote{\label{fn:video}Supplemental video: \url{https://youtu.be/pcO6XPirmUY}}
We visualize the attribution scores of an RL-based policy by continuously projecting them onto the objects that influence the robot’s decision process, visually making them glow based on their inferred importance.
Through various navigation task and obstacle configurations, we allow users to gain insights into how the robot perceives its environment and is influenced by different obstacles on its way to the goal. 
We hypothesize that this approach not only enhances the user comprehension and predictability of robot behavior, but also improves trust in the robot’s actions.


The primary contributions of our work are threefold:
\begin{itemize}
	\item A VR interface communicating robot perception and navigation policy explanations grounded in scene semantics.
    \item Extensive assessment of this novel visualization to explain robot navigation decisions in a $N=24$ user study.
    \item Empirical demonstration of significantly enhanced user understanding and predictability of the robot, with a potential for enhanced trust calibration.
\end{itemize}

\begin{figure*}[t]
	\centering
	\includegraphics[width=0.98\linewidth]{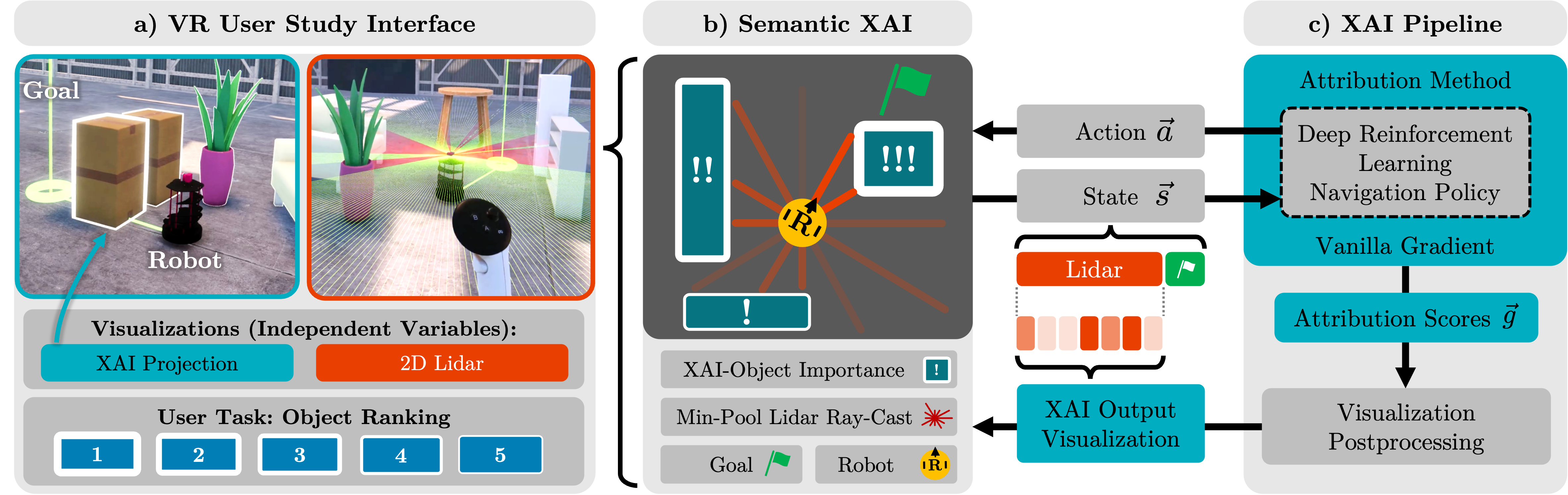}
	\caption{
        Architecture of our XAI-VR interface.
		\textbf{a)} The VR interface visualizes the robot in a navigation scenario, the object-projected XAI attribution scores, and the 2D~lidar sensor to the user.
        In our user study, the visualizations (XAI and lidar) represent the independent variables~(IVs), while we measure the users' performance in ranking the robot-surrounding objects according to their importance to the robot, defined by the visualized attribution scores.
        \textbf{b)} Objects are highlighted according to their importance by a white outline of variable thickness, here depicted in a top-down schematic.
        Their importance is assigned by the ray-casts of the 2D~lidar sensor, which project the post-processed attribution scores of the lidar-containing state space into the scene.
        Specifically, the state space contains a min-pooled set of lidar readings and the robot-centric goal position.
        \textbf{c)} The XAI technique \textit{Vanilla Gradient} generates gradient-based attributions $\vec{g}$ for the RL-trained navigation policy.
        The lidar-related part of $\vec{g}$ is post-processed for visualization in VR using Eq.~\ref{eq:postprocessing}.
\label{fig:architecture}}
\end{figure*}

\section{Related Work}
\vspace*{-0.5em}
\label{sec:related}
\subsection{General and RL-based XAI}
Explainable Artificial Intelligence (XAI) has been an active area of research for several years, leading to the development and deployment of numerous methods~\cite{speith_review_2022} across diverse application domains~\cite{tomsett_interpretable_2018}.
Core goals of XAI include enhancing transparency, trust, and human understanding of black box behavior~\cite{molnar2020interpretable}, with key questions often centered around what information should be conveyed, to whom, and in what form~\cite{liao_questioning_2020, tomsett_interpretable_2018}.
Much of the existing work in XAI has focused on standard supervised learning tasks such as classification and regression~\cite{burkart_survey_2021}.
Clearly distinct from these standard scenarios, Explainable Reinforcement Learning (XRL) has emerged as a specialized subfield aiming to bring interpretability to sequential decision-making systems trained as reinforcement-learning agents~\cite{puiutta_explainable_2020-1, milani_explainable_2024a}.
While the overarching goals and evaluation metrics of XRL align closely with those in traditional XAI, e.g., criteria such as fidelity, comprehensibility, and usefulness~\cite{nauta_anecdotal_2023, milani_explainable_2024a},the temporal nature of the RL setting and potentially complex environments motivate a clear conceptual distinction between XAI and XRL~\cite{milani_explainable_2024a}.
Milani et al.~\cite{milani_explainable_2024a} classify XRL methods into three categories based on the targeted RL agent component:
1) \textit{Feature Importance}: explaining influences on the agent in a given state,
2) \textit{Learning Process and Markov Decision Process}: identifying influential training samples, and
3) \textit{Policy-Level}: describing overall policy behavior.
We adopt a feature importance approach, using an attribution method to obtain heatmaps over sensor readings, which are subsequently processed to map sensor-level saliency to semantic scene elements for non-expert users.
Importantly, the goal of this work is not to develop a new XAI method, but rather to evaluate the effectiveness of existing XAI techniques when embedded within an immersive VR environment. 
We assess our approach through a dual evaluation strategy combining a proxy task and a questionnaire, addressing both objective performance-based and subjective user-centered criteria.
\vspace*{-0.5em}
\subsection{Explainability in Robotics and HRI}
Although XAI methods are largely developed for technical settings, applying them in human-centered robotics requires grounding abstract model outputs in ways that support user comprehension.
This is particularly important in HRI, where robotic behavior should be intuitive for users to understand.

Halilovic et al. argue for tailoring robot explanations to the users’ cognitive capabilities and task context~\cite{halilovic_robot_2024}, recognizing that overly abstract explanations may hinder comprehension among users.
In another study, they present a real-time, multimodal explanation system that incorporates robot personality and spatial context to modulate the explanation strategy~\cite{halilovic_influence_2024}.
Our interface similarly operates in real time, but focuses on visually and spatially grounding attributions of RL-based decisions through semantic object highlighting.
It therefore addresses the dynamic environmental context.

Das and Chernova introduce a framework that generates semantically grounded explanations for robot failures using scene graphs and pairwise ranking to highlight relevant spatial relations and object attributes~\cite{das_semantic-based_2021}.
Their method improves the user understanding of the robot by linking failures to specific semantic elements in the scene. 
We adopt this notion of semantic grounding and extend it to navigation, projecting attribution scores onto meaningful objects within an immersive virtual environment.

Wang et al. explore the use of augmented reality to display robot intentions to users~\cite{wang_explainable_2023}.
Their augmented reality interface aids spatial awareness and interpretability by projecting the robot's internal states into the user’s visual field. 
We adopt a similar spatial visualization paradigm but in a VR setting, enabling tighter integration of policy explanations with environmental semantics of the scene.

He et al. combine SHAP-CAM with depth-based RL to highlight influential input regions in drone navigation policies~\cite{he_explainable_2021}.
By overlaying saliency maps on depth images, they contribute a technically grounded approach to interpreting deep RL policies. 
While their visualizations remain on a technical level, our work embeds attribution-based explanations into a user-centric spatial, interactive VR interface, thereby enhancing the interpretability of RL policies through situated and dynamic visualization.

Hald et al. examine the role of robot explanations following task failures, concluding that while such explanations can guide users toward appropriate trust calibration, they are insufficient alone to repair trust~\cite{hald_error_2021}.
Rather than post hoc trust repair, our system supports continuous, real-time visual saliency explanations, aiming to proactively support the formation of calibrated trust during task execution.

Finally, Edmonds et al. investigate how different explanation modalities affect human trust in robots, comparing real-time visualizations of internal decision-making to summary text explanations~\cite{edmonds_tale_2019}.
They show that comprehensive, real-time visual feedback is more effective in fostering trust, even when not aligned with task-optimal model components. 
We adopt this insight by using dynamic visualizations of attribution scores during navigation, embedding them in a VR interface to enhance user understanding and trust.

Against this background, we hypothesize that the visualization of the XAI outputs improves 
\begin{hypothesesenum}
\item users' objective understanding of the robotic decision-making process,
\item users' subjective ability of perceiving, understanding and predicting the robotic information, and
\item calibrated trust towards the robot. 
\end{hypothesesenum}
We additionally explore whether this potential benefit is more pronounced when the visualization of XAI output is complemented with the visualization of the robot's sensor.

\vspace*{-0.5em}
\section{Methodology}
\vspace*{-0.25em}
This section introduces core concepts such as VR interface, robot navigation policy, explainability method and post-processing, and the user study setup. 

\vspace*{-0.5em}
\subsection{Virtual Reality Interface}
\label{sec:vr_interface}
To visualize a robot navigation task for the user, we develop a VR interface based on the Unity game engine, optimized for a Meta Quest 3 hardware, see Fig.~\ref{fig:architecture}.
The VR scene shows the robot navigating from a start to an end position while avoiding 3D~obstacles, e.g., furniture and other objects. 
The goal location of the robot is visualized as a green circle on the floor.
The user observes the robot navigation task from a fixed position nearby. 
Unity handles the simulation of the robot's top-mounted 2D~lidar sensor through ray-casts.
For the perceptual explainability, we visualize the otherwise invisible lidar rays in VR by rendering their 3D raycasts in real-time. 
The simulated 360~rays are displayed within the policy’s detection range of \SI{6}{\meter}. 
When a ray intersects with an object, its color changes from green to red, providing an immediate visual cue of potential obstacles.
Furthermore, 3D~objects are highlighted with an outline of dynamic width to display their importance reflecting the XAI outputs, as further elaborated in Sec.~\ref{sec:score-object-mapping}.
The Unity interface exchanges states, actions, and attribution scores with the RL policy and the attached explainability pipeline on a Python server via a socket connection.
This data is sent to the server at the inference frequency of the policy, which also triggers updates of the XAI visualizations.

\vspace*{-0.5em}
\subsection{Navigation Policy} 
\label{sec:policy}
We employ an RL-based robot navigation policy $\policy$ driven by a neural network learned using the TD3 algorithm~\cite{fujimoto_addressing_2018}.
The policy is trained for obstacle avoidance on its way to a local goal using a \SI{360}{\degree} 2D~lidar sensor in environments with randomized obstacle and goal positions.
The state $\state = [L, G]$ consists of 15 entries of min-pooled lidar sensor data $L$, down-sampled in sectors from 360 rays, and 2 entries of the robot-centric goal in polar coordinates $G$.
The policy produces a two-dimensional output dictating linear ($\velocity$) and angular ($\angle$) velocity commands for the robot as action $a_t = (v, \omega)$.
The learning task is described to the RL agent with a 
sparse goal reward ($+20$), 
sparse collision ($-20$) and 
timeout penalties ($-1$), 
jerk ($- 1\mathrm{e}{-7} \cdot\left\| (a_t - 2a_{t-1} + a_{t-2}) f^2 \right\|^2 / J_{\max}$) and 
time penalties ($-0.001$), and an 
obstacle distance-keeping penalty ($-0.001$ if $d_{\text{min}} < \SI{0.4}{\meter}$)
based on the distance to the nearest obstacle $d_{\text{min}}$.
The multi-layer perceptron policy contains three layers with [256, 128, 64] neurons respectively and is trained for $500$k time steps using Stable-Baselines3~\cite{raffin_stable-baselines3_2021}.

\vspace*{-0.5em}
\subsection{Attribution Scores of the Navigation Policy}
Attribution methods quantify the influence of each input dimension with respect to the model decision for a single input sample. Within this category, several methods have been proposed~\cite{ribeiro_why_2016, lundberg_unified_2017, sundararajan_axiomatic_2017, simonyan_deep_2014}, which differ not only in their conceptual underpinnings but some also require non-trivial choice of hyperparameters that can influence the outcome significantly~\cite{hameed_basedxai_2022}.
For its conceptual and algorithmic simplicity we use the \textit{gradient} of the policy wrt. an input state $\state$ at timestep $t$ as the attribution method~\cite{simonyan_deep_2014}, a method also known as Vanilla Gradient.
We selected Vanilla Gradient for two practical reasons: it is computationally efficient enough to support real-time inference, and it requires no hyperparameter tuning, which simplifies implementation and ensures reproducibility.
We emphasize that attribution methods explain a scalar output, i.e., in the case of our policy network $\policy$, the output of a \textit{single} neuron. 
Although explanations for both linear and angular velocity of the robot could be combined, the complexity of their interaction and the necessary communication to users exceed the scope of this work, which focuses on the VR-projection of these explanations.
Therefore, we restrict our analysis to attribution scores for the robot’s \textit{linear velocity} $\velocity$.
Further, we solely focus on explanations of the\textit{ perception-part} of the state space, hence we select the components from the gradient that correspond to the \textit{lidar components} $L$ of the input. 
The goal location $G$, while essential for task execution, serves as contextual information rather than direct sensory input and is visualized separately in the VR environment, without additional dynamic highlighting.
To summarize, the attribution scores $\grad$ are given by:
\begin{equation}
    \label{eq:gradient}
    \grad = \bigg(\frac{\partial \policy(\statet)_\velocity}{\partial \statet}\bigg)_{L}
\end{equation}

\begin{figure}[t]
	\centering
	\includegraphics[width=0.99\linewidth]{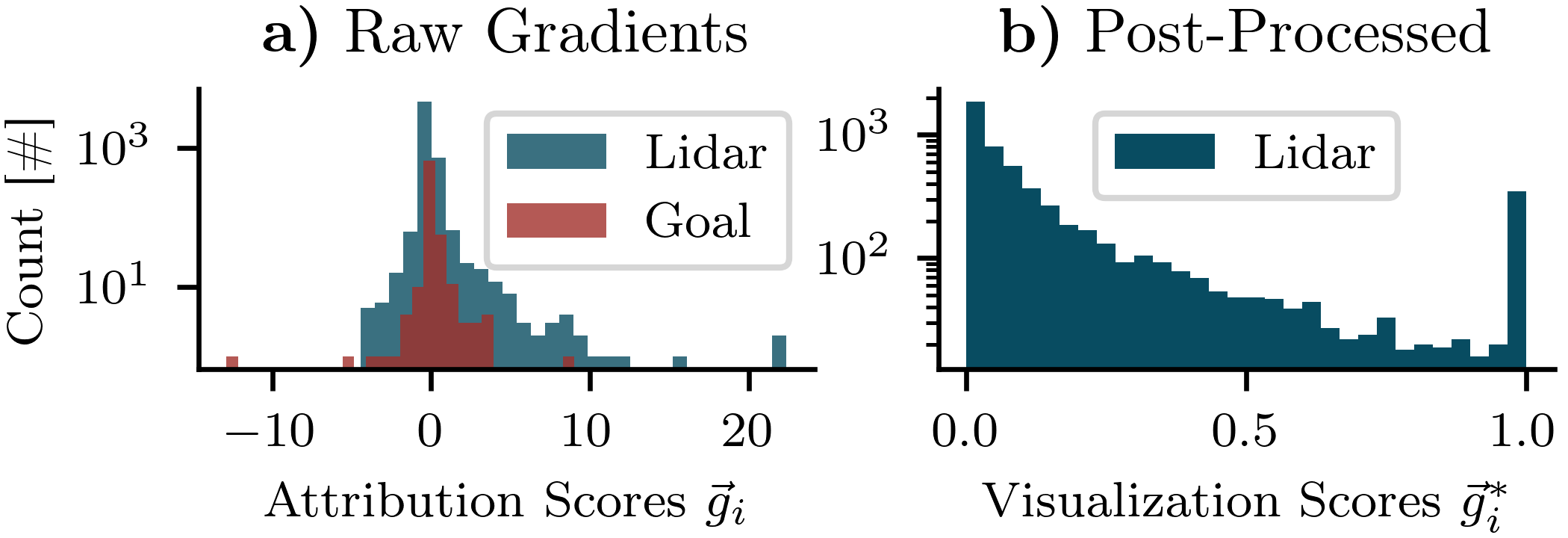}
	\caption{
        \textbf{a)} The distribution of raw lidar attribution scores $\vec{g}$ provided by Vanilla Gradient for all navigation state-action pairs presented during the user study.
        \textbf{b)} After postprocessing for visualization (Eq.~\ref{eq:postprocessing}), the distribution of $\vec{g}^*$ shifts into the range $[0,1]$.
\label{fig:attribution_score_distr}}
\end{figure}

Because the scores are derivatives, their interpretation is as follows:
Values close to zero indicate that a feature has no or little influence on the policy output. 
A high positive attribution value for a lidar input indicates that increasing the corresponding depth reading (i.e., perceiving more open space) results in a higher velocity command. 
Conversely, a high negative attribution implies that a decrease in the depth reading causes the policy to increase velocity.
\figref{fig:attribution_score_distr}a~(\textit{blue}) presents a histogram over all values in all $\grad$ provided by Vanilla Gradient for all navigation state-action pairs presented during the user study. 
Note the logarithmic scaling of the y-axis. 
The vast majority of lidar attribution scores is around zero or positive, indicating a learned tendency to reduce forward speed up upon among nearing obstacles. 
The \textit{red} histogram shows the attribution scores corresponding to the goal $G$. 
We see that scores for the goal are closer to and centered around zero, indicating that the policy primarily focuses on the lidar input.

\begin{figure}[t]
	\centering
	\includegraphics[width=0.95\linewidth]{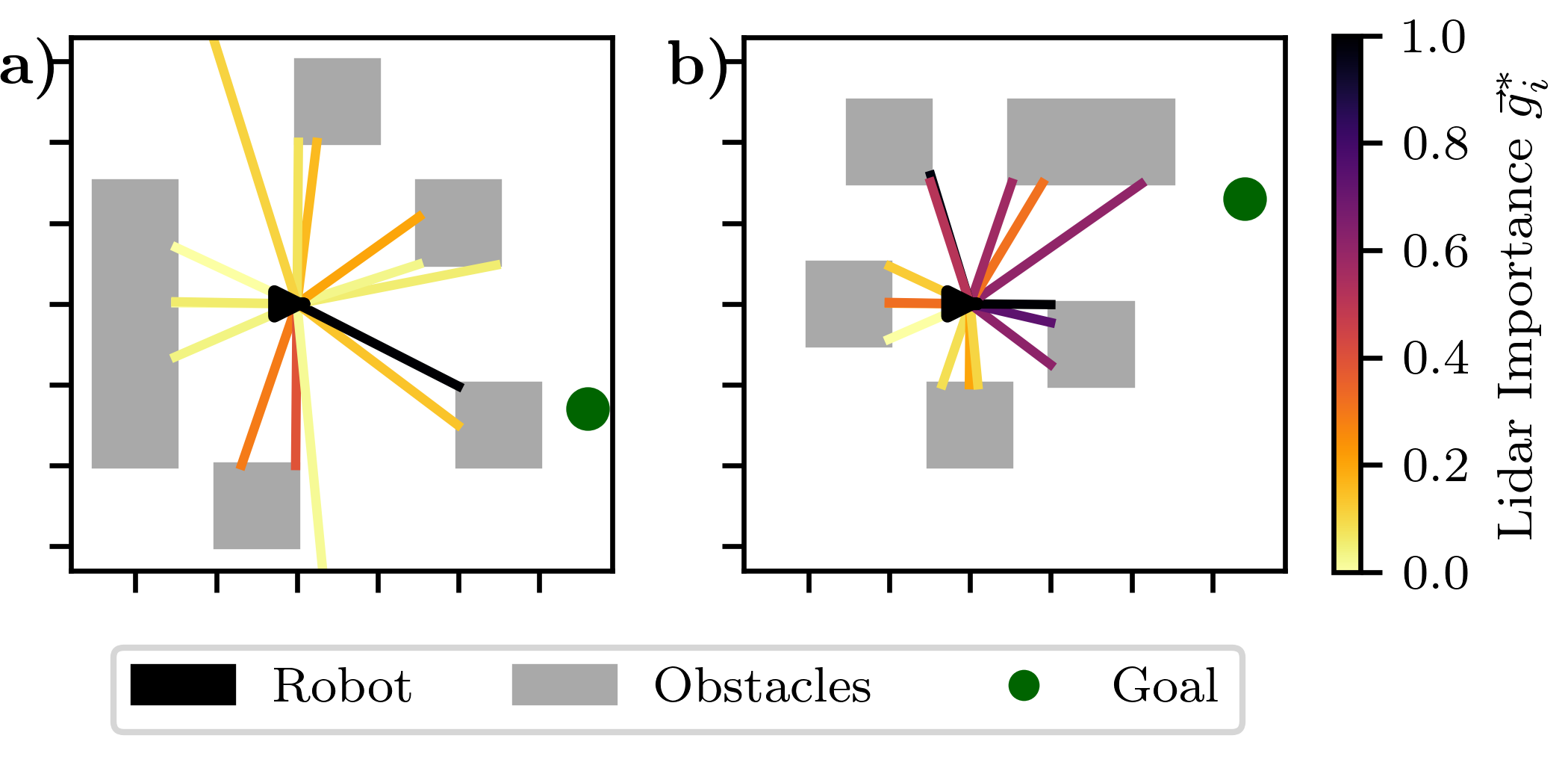}
	\caption{
        Example scenes with post-processed XAI attribution scores $\vec{g}^*$ of the linear velocity output, indicated as color-coding for their respective min-pooled lidar ray.
        The robot (black triangle) is facing to the right, while different obstacles (grey boxes)  influence the navigation policy that should pursue the goal (green dot).
        Depending on the scene setup, the obstacles influence on the policy is varies.
        Axis ticks denote $\SI{1}{\meter}$ distances.
\label{fig:xai_poc}}
\end{figure}

\vspace*{-0.6em}
\subsection{Visualizing Attribution Scores}
In order to transform the abstract numerical attribution scores into an intuitive visual representation we apply two post-processing steps: 
a) We further simplify the attribution scores and 
b) connect them with the scene by associating each with an object to achieve a semantic mapping.

\subsubsection{Simplification of Attribution Scores} 
While the \textit{sign} of the attribution scores does have a semantic meaning, as discussed above, the pure \textit{magnitude} is of far greater importance. 
Hence we work with the \textit{absolute value}, discarding the sign.
Further, the raw values of the attribution scores are less important for the ranking task than their \textit{relative} relationship, hence we apply a rescaling operation to obtain a mapping to range $[0, 1]$ for each $\grad$.
The full transformation of $\grad$ to post-processed $\grad^*$ is given by
\begin{equation}
\label{eq:postprocessing}
   \grad^* = \frac{|\grad| - \min(|\grad|)}{\max(|\grad|) - \min(|\grad|)} 
\end{equation}

The effect of this post-processing is shown in \figref{fig:attribution_score_distr}b. 
The scores now more uniformly span the full spectrum, yielding a visually uniform grading. 
The abrupt peak to the right implies that in many cases the majority of the attribution mass concentrates on few lidar rays.

\subsubsection{Score-to-Object Mapping} 
\label{sec:score-object-mapping}
In \figref{fig:xai_poc}a and b we can see an abstract visualization of the robot in two different scenes. 
The 15 lidar rays are colored according to their corresponding value in $\grad^*$. 
In \figref{fig:xai_poc}a the policy has a strong focus on the goal-occluding obstacle. 
In \figref{fig:xai_poc}b where the obstacles are generally closer to the robot, the policy's focus in the direction of the goal is less sharp. 
Overall, the backward-facing lidar rays receive less attribution.
To perform the semantic mapping of our post-processed attribution scores into the scene we associate each lidar-hit object in the robot’s vicinity with the score of the hitting lidar ray in $\grad^*$.
If an object is intersected by multiple rays, as in the illustration, the maximum value from all candidate rays is used.
Finally, object importance is visualized in the VR environment by outlining affected objects in white, using a world-space line thickness proportional to their importance.

\vspace*{-0.5em}
\subsection{User Study}
The user study is designed to evaluate the impact of different configurations of the VR interface on human objective and subjective understanding of the robot’s navigation decisions, as well as their trust in the robot.

\subsubsection{Design}
\label{sec:user_study_design}
We assess the objective understanding of the robot's navigation decisions in a ranking task, in which participants are tasked to rank the importance of objects for the robot's navigation policy in four blocks.
After each ranking block, subjective measures are taken using a questionnaire.
The questionnaire is conducted directly in VR and includes 8 questions (7-point Likert scale, labels: "Totally Agree" and "Totally Disagree"), see Sec.~\ref{sec:results_questionnaire} and Fig.~\ref{fig:survey}.

\begin{figure}[t]
	\centering
	\includegraphics[width=0.90\linewidth]{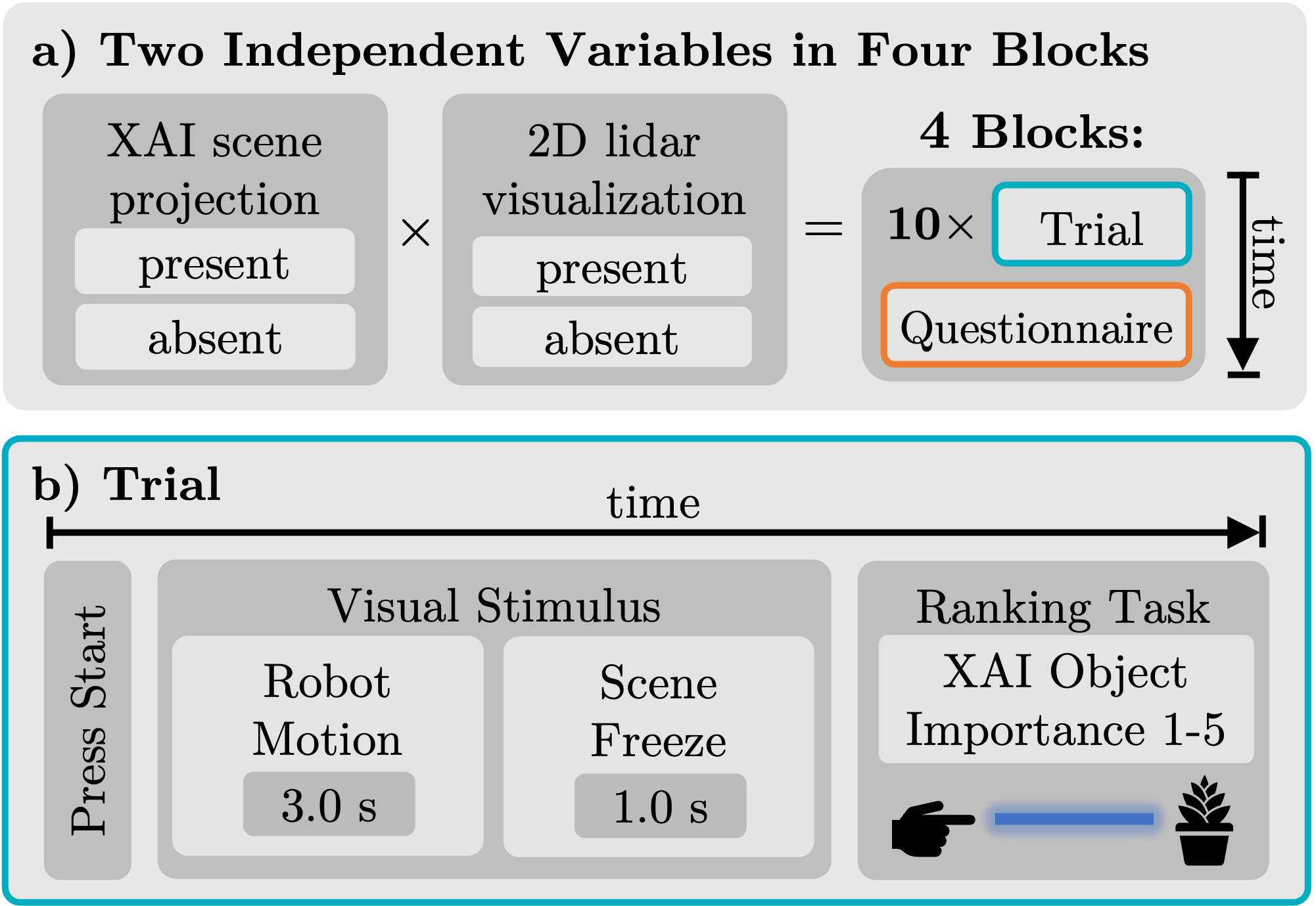}
	\caption{
        \textbf{a) }Fully-crossed combinations of two independent variables (IVs): XAI scene projection and 2D~lidar sensor visualization. 
        They sum up to four experimental conditions, represented by four blocks.
        Each block was followed by a questionnaire.
        \textbf{b) }Participants start each trial pressing the A button on the controller. 
        The robot navigated for \SI{3}{\second}, halted, and the XAI and/or lidar visualization remained after another second. 
        Afterwards, participants ranked the importance of five scene objects for the robot policy.
\label{fig:user_study}}
\end{figure}

We employ a two-factorial within-subjects design to isolate the effects of two visualization features: XAI (present, absent) and lidar (present, absent). 
Their fully-crossed combinations results in four interface configurations, as illustrated in Fig.~\ref{fig:user_study}a and video\footref{fn:video}.
Each block presents one of these configurations and consists of 12 trials with a unique navigation scenario.
Scenarios are configured by varying five obstacle placements, robot start and goal locations, and the participant's observer position.
The robot is initialized facing the goal direction.
In total, 48 unique scenarios are randomized across all four blocks.
To mitigate training and ordering effects, the sequence of the blocks is fully counterbalanced, resulting in $4! = 24$ unique orderings.
The study involves 24 participants, each assigned a different block sequence.

\subsubsection{Ranking Task}
Each trial begins with the presentation of a new robot navigation scenario including five obstacles, which the participant views from a distinct perspective, see Fig.~\ref{fig:user_study}b.
The number of obstacles is kept constant across trials to ensure a similar difficulty of the ranking task.
Depending on the experimental block, either the XAI or lidar visualization was shown. 
The robot’s goal position is indicated by a green torus, and a real-time updating line connects the goal and current robot position for a clear navigation context.
The robot starts to move when the participant presses a button of the handheld controller.
After \SI{3}{\second}, the movement is paused and marked by a stop sign on the robot. 
The final state of the visualizations remains visible for an additional \SI{1}{\second} to allow the participant to process the current navigation step, which they are instructed to base their ranking on.

Participants then rank the importance of each object with respect to the robot’s policy by pointing and selecting the objects. 
Rank labels are displayed on top of the objects, ordered from most (1) to least (5) important. 
Participants can revise their ranking decision by pressing another button.

The collected rankings are later compared to ground-truth object importance derived from scene-projected attribution scores.
To measure agreement between the participant’s object ranking and the ground-truth importance order, we employ Kendall’s $\tau$~\cite{kendall_new_1938}, a non-parametric correlation metric. 
Kendall’s $\tau$ quantifies similarity between rankings by evaluating the proportion of concordant and discordant pairs.

\subsubsection{Procedure}
Before the experiment, participants received detailed instructions about the experiment, provided written consent, and completed a demographic questionnaire.
They were informed about the robot’s navigation task, the XAI output visualization and how its lidar sensor perceives the environment.
The experimenter instructed them for the ranking tasks (S1).
Each participant completed two training trials with explanations to become familiar with the visualizations and ranking task, and proceeded with the first experimental block.
After they had completed the first ranking block, they answered the questionnaire measuring the subjective experience of the previously presented interface configuration (S2).
Upon completion of all ranking blocks and questionnaires, participants answered a freeform questionnaire targeting their object ranking strategy (S3).

\subsubsection{Participants}
A total of $N=24$  individuals (9 women, 15 men) participated in the study in exchange for a EUR 10 monetary compensation.
All participants reported having (corrected-to-) normal vision.
Their mean age of was 24.6 years (\textit{SD} = 3.6 years).
Participants rated their experience with AR/VR on a 7-point Likert scale (1 = No experience at all, 7 = A lot of experience), with a mean rating of 2.4 (\textit{SD} = 1.3).
Participants also rated their experience with robotics (\textit{M} =2.8, \textit{SD} = 1.9), and their experience with artificial intelligence (\textit{M} = 4.1, \textit{SD} = 1.8).
The study adhered to the principles outlined in the Declaration of Helsinki.

\vspace*{-0.5em}
\section{Experimental Evaluation}
\vspace*{-0.25em}
\label{sec:exp}
This section presents the results of the user study, which evaluates the established hypotheses (H1 - H3).

\vspace*{-0.5em}
\subsection{User Study}
The collected data covers the objective visualization-dependent object ranking performance (S1) and subjective evaluations of the post-block questionnaire (S2).

\subsubsection{Ranking}
\begin{figure}[t]
	\centering
	\includegraphics[width=0.98\linewidth]{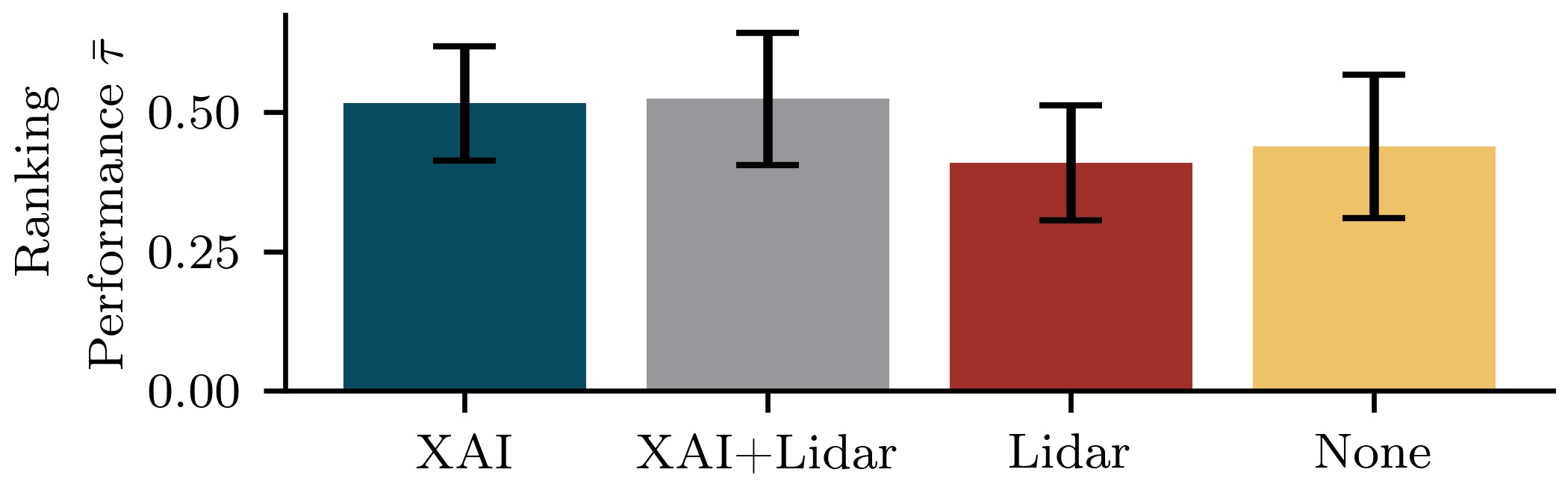}
	\caption{
        Object ranking performance (S1) of participants for each presented visualization combination of XAI and lidar conditions as measured by Kendall's $\tau$. 
        Means and standard deviations are shown.
        As can be seen, the XAI visualizations increase the users' object ranking performance with respect to their attribution score-derived ground truth importance.
\label{fig:ranking_performance}}
\end{figure}

\begin{table}[t]
\centering
\begin{tabular}{rccccc}
\hline
\textit{\textbf{Predictor}} & $df_d$ & $df_{n}$ & $F$ & $p$ & $\eta^2_p$ \\
\hline
\textbf{XAI} & \textbf{1} & \textbf{23} & \textbf{15.20} & \textbf{$<$ .001} & \textbf{0.40} \\
Lidar & 1 & 23 & 0.20 & .657 & 0.01 \\
XAI \( \times \) Lidar & 1 & 23 & 0.92 & .346 & 0.04 \\
\hline
\end{tabular}
\caption{The results of rmANOVA of the ranking task performance (S1) demonstrate a significant effect of the semantic XAI visualization.}
\label{tab:rm_anova}
\end{table}

The ranking task (S1) quantitatively assessed users’ understanding of the XAI visualizations. 
We compute Kendall's $\tau$ between the participants' ranking and the ground truth order of objects for every trial and aggregate the results for each of the four experimental conditions and each participant, see Fig.~\ref{fig:ranking_performance}.

A repeated-measures (rm)ANOVA confirms a significant effect of the XAI visualization on the participants' ranking performance of the five scene objects, see Table~\ref{tab:rm_anova}.
Participants performed better with ($M=0.52$, $SD=0.11$) than without XAI ($M=0.42$, $SD=0.12$).
This benefit is expected as the XAI visualization conveys attribution scores.
Although neither the main effect of the lidar visualization nor its interaction with the XAI visualization was significant, participants achieved descriptively the best ranking performance, when both XAI and lidar were visible.

Interestingly, even without visualizations, participants achieved a certain ranking accuracy, possibly by using heuristics, e.g., prioritizing objects closer to or in front of the robot, or those that appeared to influence its navigation.

While edge cases such as ties in the ground truth order due to occluded objects receiving a zero importance score or subtle importance differences indistinguishable from outline thickness can impact Kendall’s $\tau$ as the absolute ranking performance, the relevant conclusions are to be drawn from the relative performance differences. 

In conclusion, the participants showed an improved understanding of the object importance to the RL policy with the XAI visualization, finding support for H1.  

\subsubsection{Questionnaire}
\label{sec:results_questionnaire}
\begin{figure}[t]
	\centering
	\includegraphics[width=0.95\linewidth]{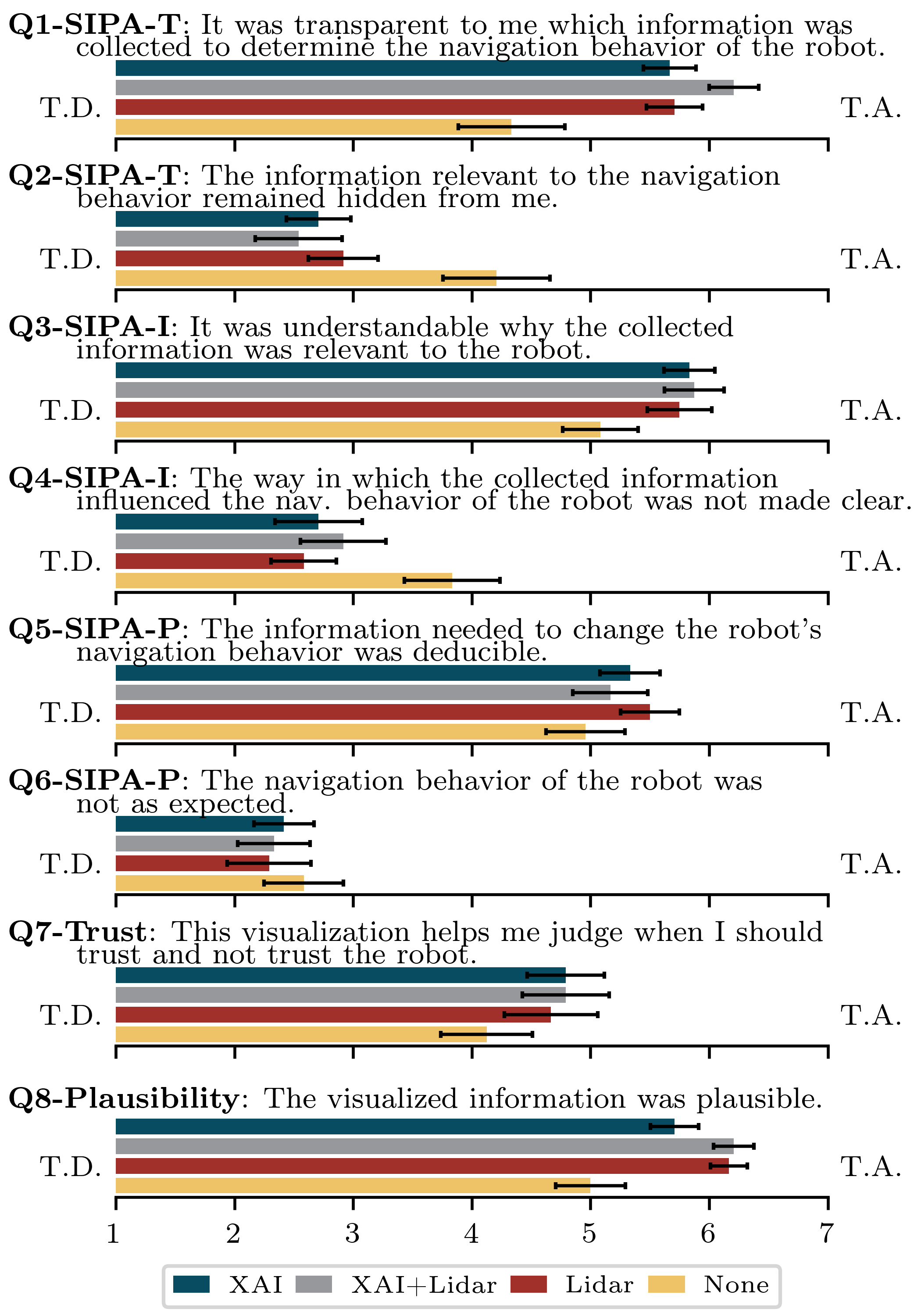}
	\caption{
    Participants rated their experience with respect to the SIPA scale for explanations (Q1-Q6), trust (Q7) and plausibility (Q8) with regards to the visualizations XAI and lidar (IVs).
    All questions shared the same labels: Totally disagree (T.D.) and Totally Agree (T.A.), abbreviated here for visual clarity.
    Ratings were provided on a Likert scale (1-7), bars indicate score means, error bars show standard errors.
    \label{fig:survey}
        }
\end{figure}

\begin{table}[t]
\centering
\begin{tabular}{lrccccc}
\hline
\textit{\textbf{Item/Scale}} & \textit{\textbf{Predictor}} & $df_d$ & $df_{n}$ & $F$ & $p$ & $\eta^2_p$ \\
\hline
\multirow{3}{*}{Q1-6: SIPA} & \textbf{XAI} & \textbf{1} & \textbf{23} & \textbf{5.39} & \textbf{.030} & \textbf{0.19} \\
 & \textbf{Lidar} & \textbf{1} & \textbf{23} & \textbf{4.43} & \textbf{.046} & \textbf{0.16} \\
 & \textbf{XAI \( \times \) Lidar} & \textbf{1} & \textbf{23} & \textbf{6.90} & \textbf{.015} & \textbf{0.23} \\
\hline
\multirow{3}{*}{Q7: Trust} & XAI & 1 & 23 & 1.62 & .216 & 0.07 \\
 & Lidar & 1 & 23 & 1.72 & .202 & 0.07 \\
 & XAI \( \times \) Lidar & 1 & 23 & 0.80 & .380 & 0.03 \\
\hline
\multirow{3}{*}{Q8: Plaus.} & XAI & 1 & 23 & 3.86 & .062 & 0.14 \\
 & \textbf{Lidar} & \textbf{1} & \textbf{23} & \textbf{22.77} & \textbf{$<$ .001} & \textbf{0.50} \\
 & XAI \( \times \) Lidar & 1 & 23 & 3.26 & .084 & 0.12 \\
\hline
\end{tabular}
\caption{Results of the rmANOVAs for post-block questionnaire (S2), specifically for the short SIPA scale (Q1-Q6), Q7-Trust and Q8-Plausibility.}
\label{tab:rm_anova_questionnaire}
\end{table}

Targeting the users' understanding of our explanation visualizations in terms of transparency (T), intelligibility (I), robot predictability (P), and trust (T) towards the robot, we analyze the 8-item questionnaire (Likert scale, score 1-7) of S2, see \textbf{Fig.~\ref{fig:survey}}.
Reverse-coded items (Q2, Q4, Q6) served as attention checks.

Items Q1-Q6 represent a short version of the Subjective Information Processing Awareness (SIPA) scale~\cite{schrills_are_2021} for user-centered assessment of XAI: Perception (Q1+2), intelligibility (Q3+4) and prediction (Q5+6).
We invert the reverse-coded items, aggregate the scores of the single SIPA items to a mean score for each experimental conditions, and perform a two-factorial rmANOVA (Tab.~\ref{tab:rm_anova_questionnaire}) to infer the contributions of the XAI and lidar visualization conditions.

Both XAI and the interaction of XAI and lidar show a significant effect on the SIPA score, with XAI ($M=5.5$, $SD=1.0$), XAI+Lidar ($M=5.6$, $SD=1.0$), and Lidar only ($M=5.5$, $SD=1.0$) achieving higher SIPA scores compared to the condition without XAI and lidar ("None") ($M=4.6$, $SD=1.4$).
This underlines that any additional visualization of the robot's information processing (XAI, lidar or both) improves the participants' impression of being able to perceive, understand and predict the navigation behavior, supporting H2.

Item Q7 assesses participants' trust calibration towards the robot and was derived from the Explanations Satisfaction Scale (ESS)~\cite{hoffman_measures_2023}.
While visualizations (XAI, XAI+lidar, lidar) descriptively outperformed “None,” no significant effects were found (not supp. H3).
This absence of significance may be related to specifics of the study design:
First, the single-item metric may have lacked sensitivity to detect changes in trust calibration.
Second, the visualizations may have been irrelevant to trust calibration.
Finally, participants observed error-free robot navigation from a distance without personal involvement, which may have obscured what exactly they were expected to trust the robot for.

The final item Q8 targets the plausibility of the visualized information and refers to the explanation concept of coherence~\cite{nauta_anecdotal_2023}.
Here, the lidar has a significant effect on the measured plausibility of visualizations.
Both XAI and its interaction with lidar are not significant.
This indicates that the lidar visualization rather than the XAI visualization appears more plausible for users, presumably because the lidar rays are directly linked to the robot's perception.

We conclude that the semantic XAI projection helped the users to objectively perceive the information leveraged by the navigation policy, and also created the subjective impression for users of being able to perceive, understand and predict the robot's information processing, i.e., decision-making in navigation behavior (supporting H2).
While the objective understanding in the ranking task was not significantly affected through the visualization of the lidar rays, the subjective information processing awareness (perception, understanding, prediction) of the users as well as the perceived plausibility of the interface improves.
Finally, neither the XAI projection nor the lidar visualization changes the user's impression of enhancing the trust calibration process.

\subsubsection{Freeform Feedback}
Upon completion of all blocks, we asked participants two freeform questions (S3) to learn about their mental model (RQ3) and object ranking strategy:
FQ1 - What rules do you think the robot followed when choosing a path?
FQ2 - Which strategy did you use for ranking objects' influence?
Regarding robot path selection rules (FQ1), participants frequently stated that the robot prioritized collision avoidance, selected shorter and direct routes for efficiency, showed differential treatment to objects based on their distance, and favored smooth trajectories. 
Almost all responses built upon the constellation of objects, i.e., the scene context, rather than the perception and action capabilities of the robot.
This underlines the relevance of scene context for the participants' mental model of explainability and the need to promote their awareness of the robot's perception.
For the ranking task (FQ2), common strategies included considering outline thickness, object size or perceived collision hazard, proximity to the robot’s intended path, and frequently a combination of these factors.

\vspace*{-0.75em}
\section{Conclusion}
\vspace*{-0.25em}
\label{sec:conclusion}
We present a novel VR-based interface that integrates dynamic, scene-grounded XAI outputs and sensor visualizations to support non-expert users in understanding an RL-based robot navigation policy.
We thereby align numerically obscure robot policy explanations to the users’ cognitive capabilities and task context.
Our user study shows that semantically projecting attribution scores significantly improves non-expert users’ objective understanding and subjective awareness of the robot’s decision-making, thereby increasing the perceived predictability of its behavior.
Visualizing the robot’s lidar rays also contributes substantially to users’ subjective awareness, indicating that combining XAI and sensor visualizations is essential for optimizing user experience in VR.
Based on these findings, future research should jointly evaluate objective and subjective metrics to guide the design of effective explanation tools for HRI.
A possible future direction is incorporating the temporal dimension of robot trajectories into the XAI method, particularly in environments with dynamic obstacles, and how users generalize acquired understanding to different navigation scenarios.
Overall, our results highlight the potential of immersive VR explanation interfaces to facilitate more transparent human-robot interaction in complex environments, with applications in supporting safer collaboration, aiding debugging and validation, and facilitating training and education.
\bibliographystyle{IEEEtran}
\vspace*{-0.50em}
\bibliography{bib_xai_complete, bib_sebastian}
\end{document}